\documentclass[manuscript=article]{achemso}

\usepackage{amsmath}

\usepackage[utf8]{inputenc} 
\usepackage[T1]{fontenc}    
\usepackage{hyperref}       
\usepackage{url}            
\usepackage{booktabs}       
\usepackage{amsfonts}       
\usepackage{nicefrac}       
\usepackage{microtype}      
\usepackage[table]{xcolor}         
\usepackage{graphicx}
\usepackage{multirow}
\usepackage{float}
\usepackage{amssymb}
\usepackage{array}
\usepackage{subcaption}
\usepackage{mathtools} 
\usepackage{nicematrix}
\usepackage{tablefootnote}
\usepackage{natbib}
\usepackage{subcaption}
\usepackage{standalone}
\usepackage{amsmath}
\usepackage{siunitx}
\usepackage{algorithm}
\usepackage{algpseudocode}
\usepackage{import}
\usepackage[draft]{minted}
\usepackage{siunitx}
\usepackage{listings}
\usepackage{amssymb}
\graphicspath{ {./figures/} }

\author{Dule Shu}
\affiliation[MechE]{Department of Mechanical Engineering,
Carnegie Mellon University,
Pittsburgh PA, USA}
\altaffiliation{Contributed equally to this work}
\author{Wilson Zhen}
\affiliation[MechE]{Department of Mechanical Engineering,
Carnegie Mellon University,
Pittsburgh PA, USA}
\altaffiliation{Contributed equally to this work}
\author{Zijie Li}
\affiliation[MechE]{Department of Mechanical Engineering,
Carnegie Mellon University,
Pittsburgh PA, USA}
\author{Amir Barati Farimani}
\affiliation[MechE]{Department of Mechanical Engineering,
Carnegie Mellon University,
Pittsburgh PA, USA}
\alsoaffiliation[ML]{Machine Learning Department,
Carnegie Mellon University,
Pittsburgh PA, USA}
\alsoaffiliation[Chem]{Department of Chemical Engineering,
Carnegie Mellon University,
Pittsburgh PA, USA}
\email{barati@cmu.edu}

\title{Inpainting Computational Fluid Dynamics with Deep Learning}

\begin{document}
\newgeometry{margin=1in}
\begin{abstract}
Fluid data completion is a research problem with high potential benefit for both experimental and computational fluid dynamics. An effective fluid data completion method reduces the required number of sensors in a fluid dynamics experiment, and allows a coarser and more adaptive mesh for a Computational Fluid Dynamics (CFD) simulation. However, the ill-posed nature of the fluid data completion problem makes it prohibitively difficult to obtain a theoretical solution and presents high numerical uncertainty and instability for a data-driven approach (\textit{e.g.}, a neural network model). To address these challenges, we leverage recent advancements in computer vision, employing the vector quantization technique to map both complete and incomplete fluid data spaces onto discrete-valued lower-dimensional representations via a two-stage learning procedure. We demonstrated the effectiveness of our approach on Kolmogorov flow data (Reynolds number: 1000) occluded by masks of different size and arrangement. Experimental results show that our proposed model consistently outperforms benchmark models under different occlusion settings in terms of point-wise reconstruction accuracy as well as turbulent energy spectrum and vorticity distribution.

\end{abstract}
\restoregeometry
\section{Introduction}
High-fidelity fluid dynamics data from physical experiments or numerical simulation is indispensable for precise analysis and decision-making across many scientific and engineering domains. However, acquiring such data is often challenging due to limitations on available resources, such as the restricted placement and resolution of measurement devices, and the budget for computational software/hardware. The demand for high-fidelity fluid dynamics data has spurred the development of various data reconstruction and assimilation techniques \citep{habibi2021integrating, liu2022new, yousif2021high, fu2023semi, shu2023physics}. In this work, we are interested in reconstructing full field flow data given incomplete observations. More specifically, we assume that the missing part in the field is large enough such that it affects the flow structures of all scales and limited clues can be inferred from the nearby region, therefore a simple interpolation is likely to fail because of the large gap between the available context. 

The main challenge of fluid flow reconstruction is its ill-defined nature. Let $\mathcal{D}:=\{x_1,x_2,\cdots,x_k\}$ be a set of collocation points where $\omega_t$, the numerical solution of a Partial Differential Equation (PDE) at timestep $t$, is evaluated. In a data completion problem setting, the discrete domain $\mathcal{D}$ is divided into a region $\mathcal{D}^{\text{cond}}$ where the value of $\omega_t$ is known and a region $\mathcal{D}^{\text{mask}}$ where the value of $\omega_t$ is unknown (masked out), the goal of data completion is to find the function values $\{\omega_t \left(x_i\right)|x_i \in \mathcal{D}^{\text{mask}}\}$ given the available values $\{\omega_t \left(x_j\right)|x_j \in \mathcal{D}^{\text{cond}}\}$ as the conditioning information. In comparison, a typical numerical solver obtains $\{\omega_t \left(x_k\right)|x_k \in \mathcal{D}^{\text{mask}}\}$ using the initial condition $\omega_0$ and the boundary conditions $\partial\omega_1, \partial\omega_2, \cdots, \partial\omega_t$ from $t$ timesteps. Since $\{\omega_0, \partial\omega_1, \cdots, \partial\omega_t\}$ in general contains more information than $\{\omega_t \left(x_j\right)|x_j \in \mathcal{D}^{\text{cond}}\}$, a data completion method does not have sufficient information to obtain a unique accurate solution as a numerical solver has. Therefore, with the data completion method, the best one can hope for is to find a mapping $f: \mathbf{\omega_t}\left(\mathcal{D}^{\text{cond}}\right)\rightarrow \mathbf{\omega_t}\left(\mathcal{D}^{\text{mask}}\right)$ such that the error in predicting $\{\omega_t \left(x_i\right)|x_i \in D^{\text{mask}}\}$ from $\{\omega_t \left(x_j\right)|x_j \in \mathcal{D}^{\text{cond}}\}$ is minimized. In this paper, we are interested in assessing how deep neural networks can perform as the function approximator of $f$. In particular, we choose 2D Kolmogorov flow at a resolution of $256 \times 256$ obtained from a numerical solver as the ground truth data for model evaluation, as its conspicuous pattern of turbulence poses more challenge to data completion and offers us a perspective to observe how the property of vortex and the choice of masks will affect the reconstruction accuracy.

\begin{figure}[t]
    \includegraphics[width=\linewidth]{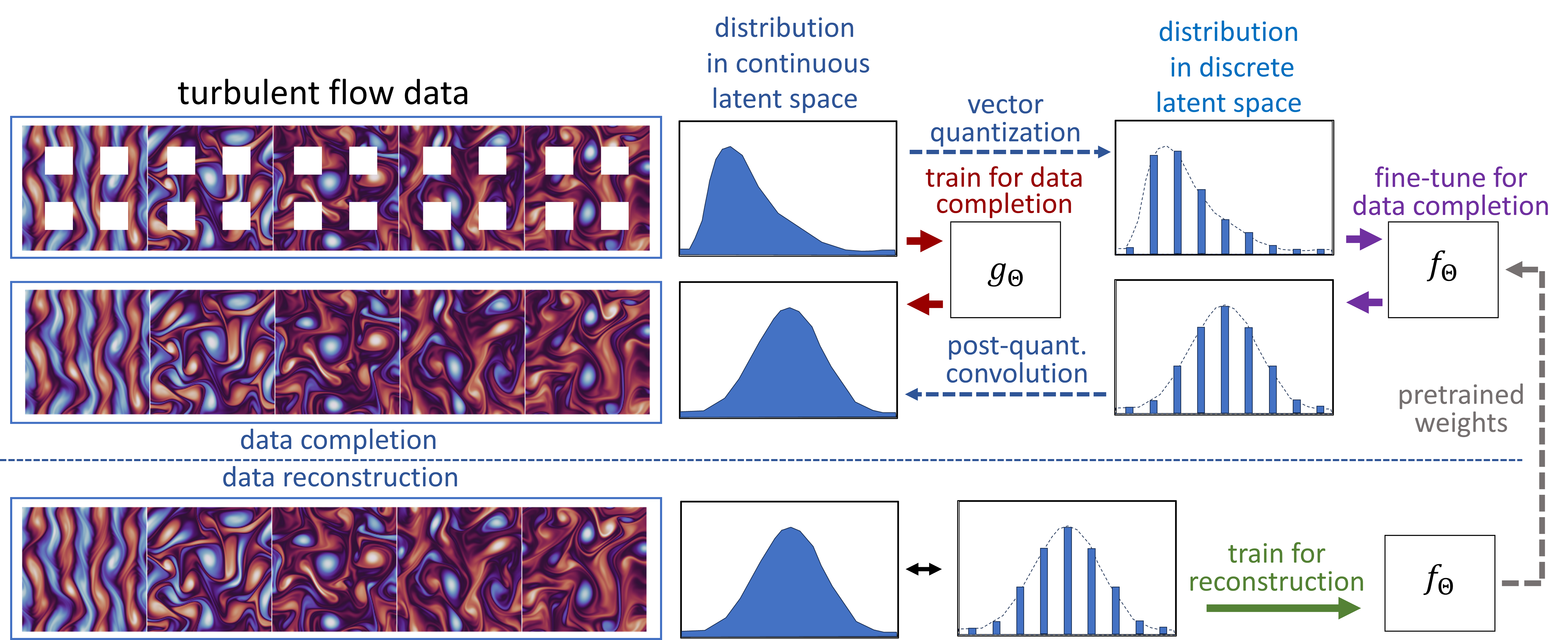}
    \captionof{figure}{An overview of our two-stage method for data completion. Our model (denoted as $f_{\Theta}$) is trained for data construction in the first stage to learn a latent representation of the complete data in a latent space the discretized by vector quantization (as shown in the lower portion of the figure). The model is then fine-tuned for data completion task in the second stage, with a post-quantization convolution module applied to the prediction in the discrete latent space to reduce the artifact caused by quantization. In comparison, a model without a VQ module (denoted as $g_{\Theta}$) directly learns to predict complete data from incomplete input in some continuous latent space (as shown in the upper portion of the figure). \label{fig:method-overview}}
\end{figure}%

The recent progress in the field of image synthesis with deep neural networks \citep{zhu2020unpaired, goodfellow2014generative, karras2019stylebased, ho2020denoising} has opened up various new approaches for fluid field analysis \citep{li2020learning, pant2021deep, howard2023multifidelity, hemmasian2023multi, lorsung2024picl, lorsung2024physics}. One of these applications aim to use neural networks as learned interpolation to super-resolve the field information \citep{fukami2019super, gao2021, yousif2021high, fukami2021machine, kim2021unsupervised, Shu2023pi-diff, pant2021deep, belbuteperes2020combining, li2021tpu, xie2018tempogan, esmaeilzadeh2020meshfreeflownet, ren2023physr}. The super-resolved results can also be combined with numerical solver to reduce the error arising in under-resolved grid \citep{um2020solver, kochkov2021machine, sun2023neural}. Many of these neural network models require the stencils in the input being uniformly scattered and only the local details between stencils are under-resolved. Another direction is to use neural networks to predict the state of the system \citep{PINO_li_2021, li2022transformer, li2022graph, dl-rom2021, Thuerey2020rans, latentphysics2021, Sun2020surrogate, mp-pde-solver, pfaff2021learning, Jin2021nsfnet}, thus bypassing the process of calling numerical solver. However, it is commonly observed that neural networks tend to produce unbounded error accumulation on time-dependent system \citep{mp-pde-solver, pfaff2021learning} and a stable long-term prediction is generally difficult. The data completion task can be seen as the intersection of the two aforementioned applications, where the neural networks learn to interpolate the smaller scale details while also predicting the entire field. One line of works on fluid data completion involves using principal component analysis to obtain a set of the bases of the collected data and fitting the corresponding coefficients for the field of interest \citep{saini2016development, venturi2004gappy}, which relies on the linearity assumption in the reduced space. More recently, deep neural networks have been shown to be a plausible choice for data-driven fluid flow completion \cite{pathak2016inpaint, song2016semantic, Yu_2018_CVPR}. \citet{Buzzicotti2021inpaint} propose a generative adversarial network framework for completing the missing region in 3D rotational turbulence. \citet{foucher2020physicsaware} propose a framework based on U-Net \citep{ronneberger2015unet} to predict the missing data with a loss function based on the residual of governing equation. 
Note that the problem of fluid flow completion is connected but different from the long-studied problem of image inpainting in the computer vision community, as the main purpose of image inpainting is to produce a visually plausible prediction for the missing region of an image, whereas the goal of fluid flow completion is bond to a unique ground truth reference of the missing region from either the solution of the governing PDE or the fluid dynamics experiment data. Nevertheless, many methods used in inpainting deep learning models are potentially applicable to designing the deep learning tool for fluid flow completion, including progressive inpainting \citep{li2020recurrent}, usage of structural information (\textit{e.g.}, gradient) as guidance \citep{yang2020learning}, combining adversarial loss and/or perceptual loss with reconstruction loss \citep{johnson2016perceptual, song2018spg}, and a GAN inversion structure \citep{pan2021exploiting}.

In this work, we propose to solve the fluid flow data completion problem with a two-stage procedure. In the first stage, we use an auto-encoding neural network model to learn a latent representation of the original 2D turbulence data. In the second stage, we fine-tune the encoding module from the first stage to predict the unmasked data samples in their latent form, while evaluating the predicting error using the fixed-weight decoder trained in the first stage.
To exploit the low-dimensional structure in the data, we use vector quantization (VQ) technique to learn a discrete latent space \citep{oord2018neural}, where the original data is encoded as the composition of a set of vector bases. Compared to the continuous autoencoder regularized with Gaussian prior (also known as Variational Auto-Encoder, VAE) \cite{kingma2022autoencoding}, VQ-VAE has been empirically shown to have better sampling quality \citep{oord2018neural, razavi2019generating}. An overview of our proposed two-stage method for fluid flow data completion is illustrated in Fig. \ref{fig:method-overview}. 
We chose to benchmark our data completion method with two representative works from the field of neural network-based operator learning, the Fourier Neural Operator (FNO) \cite{li2020fourier} and the Factorized Transformer (FactFormer) \cite{li2024scalable}. Neural operator \citep{cao2021choose, li2023latent, patil2023hyena, Goswami2023} is a class of neural network models designed to learn the mapping between two function spaces specified by a PDE, \textit{e.g.}, the mapping from the initial condition to the solution of the PDE at a given time as in the models' popular application of Initial Value Problems (IVP). To learn such mapping, a neural operator typically maps the input data to some latent space (or channel space as referred to in FNO) before computing the mapping as a kernel integral of the latent variable parameterized by neural network weights (For example, in FNO, the kernel integral is computed in the Fourier space, while in FactFormer, the kernel integral is computed with the Transformer's attention mechanism one data-dimension at a time). Theoretically, neural operators can be directly used to solve the fluid flow data completion problem, as the mapping from initial condition to solution at target time in VIP can simply be substituted by the mapping from the unmasked region $\mathbf{\omega_t}\left(\mathcal{D}^{\text{cond}}\right)$ to the masked region $\mathbf{\omega_t}\left(\mathcal{D}^{\text{mask}}\right)$. Through numerical experiments, however, we found out that both neural operator benchmarks with a continuous latent space are outperformed by our proposed method which learns the mapping in the discrete latent space via vector-quantization, a result possibly due to the ill-defined nature of data completion. To the best of our knowledge, our method is the first work that generates competitive results for 2D turbulent flow data completion at the resolution of $256\times 256$, whereas the most similar work to ours by \citet{Buzzicotti2021inpaint} is proposed for the data resolution of $64\times 64$ and without the goal towards optimized performance as explained by its authors.

\section{Method}
\subsection{Problem Formulation}
Let $\mathbf{X}:=\left(X, X_\text{mask}, B\right)\in\mathcal{R}^{3\times H\times W}$ be a data sample of three components defined for a data completion problem. The first component, $X\in\mathcal{R}^{H\times W}$, denotes the vorticity of the fluid flow evaluated on a 2D grid of height $H$ and width $W$ (referred to as the ground truth data), the second component, $X_\text{mask}\in\mathcal{R}^{H\times W}$, denotes the vorticity evaluated on a part of the grid with missing values replaced by zeros (referred to as the masked or incomplete data), and the third component $B\in{\{0,1\}}^{H\times W}$ (referred to as the mask) is a Boolean matrix specifying which region(s) of the grid have vorticity values masked out, with 0 indicating a node with missing value and 1 indicating a node with known value. The problem of fluid flow data completion is then formulated as finding a function $f_\Theta$ parameterized by $\Theta$ (\textit{e.g.}, the set of neural network weights) such that $f_\Theta \left(X_\text{mask}, B\right)$ approximates $X$. By optimizing $\Theta$ over a training dataset, we aim to accurately reconstruct $X$ from $\left(X_\text{mask}, B\right)$ on a test dataset.

\begin{figure}[t]
    \includegraphics[width=\linewidth]{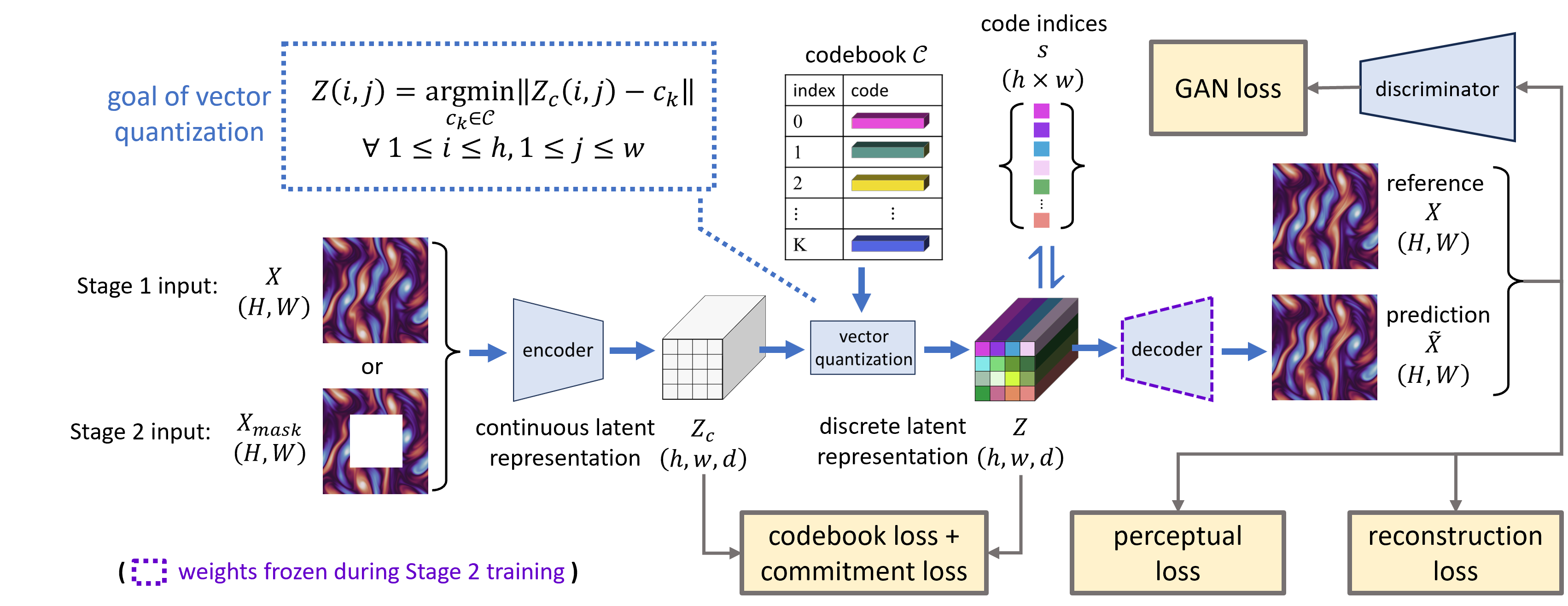}
    \captionof{figure}{Model architecture and training procedure of VQ-VAE for data completion. The major components of a VQ-VAE model includes an encoder, a decoder, a vector quantization module and the codebook associated with it, and a discriminator to implement GAN loss. The decoder module is trainable during Stage 1 and frozen during Stage 2 such that the data completion is eventually performed in the VQ space. \label{fig:training_procedure}}
\end{figure}

The procedure to obtain $f_\Theta$ consists of two stages. In Stage 1, an auto-encoding neural network model is trained for data reconstruction in order to learn a latent representation of $X$. In Stage 2, model fine-tuning is applied to the weights obtained from the previous stage, with the goal switched from data reconstruction to data completion, and the model input correspondingly switched from complete data sample $X$ to a combination of masked data sample $X_\text{mask}$ and the mask $B$. To address the numerical uncertainty and instability in prediction due to the ill-posed nature of the data completion problem, a Vector-Quantized Auto-Encoder (VQ-VAE) is chosen as the backbone model for the two stages. The VQ-VAE model defines a discrete-valued latent space such that each value can be represented by an integer code from a learned codebook. Such discretization helps to stabilize the data prediction task in Stage 2. The major components of the VQ-VAE architecture and model training are described in Fig. \ref{fig:training_procedure}. The More details of model design and the associated two-stage learning are provided in the following two subsections.

\subsection{Codebook and Autoencoder Learning (Stage 1)}
The goal of Stage-1 learning is to find a latent representation of $X$. Our VQ-VAE model follows the design by \citet{oord2018neural} which comprises three main modules, an encoder, a decoder, and a vector-quantization module. Given a complete data sample $X\in\mathcal{R}^{H\times W}$ of fluid flow, the encoder $E$ computes an embedding $Z_c=E\left(X\right)\in \mathcal{R}^{h\times w\times d}$. The vector-quantization module $q$ then converts $Z_c$ to its quantized counterpart $Z=q\left(Z_c\right)\in \mathcal{R}^{h\times w\times d}$. Let $Z_{c} \left(i,j\right)$ ($1\leq i \leq h, 1\leq j \leq w$) denote the $d$-dimensional vector extracted from $Z_c$ at its $i-$th position in the first dimension and its $j-$th position in the second dimension, and let $Z\left(i,j\right)$ be defined similarly from $Z$. $q$ maps $Z_c \left(i,j\right)$ to its nearest neighbor from a learnable codebook $\mathcal{C}:=\{c_k \in \mathcal{R}^d \}_{k=0}^K$ by solving the following optimization problem.
\begin{equation}\label{eq:vector quantization}
    \begin{gathered}
        Z \left(i,j\right) = \operatorname*{argmin}_{c_k \in \mathcal{C}}\|Z_c \left(i,j\right)-c_k\|_2
    \end{gathered}
\end{equation}
Finally, the decoder $D$ is used to obtain $\Tilde{X}:=D\left(Z\right)\in\mathcal{R}^{ H\times W}$, a reconstruction of $X$. Together, $E$, $q$ and $D$ constitute the VQ-VAE model $f_\text{rec}:=D \circ q \circ E$ for learning the latent representation of $X$. In order to obtain a perceptually rich VQ-VAE model for data reconstruction, we adopt the training strategy for VQ-GAN \cite{esser2021taming} which combines the standard VQ-VAE loss function with GAN loss and perceptual loss. The VQ-GAN loss function, denoted as $\mathcal{L}_{VQ}$, consists of three components: the reconstruction loss, the codebook loss and the commitment loss. Formally, $\mathcal{L}^{VQ}$ is defined as follows.
\begin{equation}\label{eq:stage_1_loss}
    \mathcal{L}_{VQ}={\Vert{X-\Tilde{X}}\Vert}^2_2 + \Vert{sg\left(Z_c\right)-Z}\Vert^2_2+\beta\cdot\Vert{Z_c-sg\left(Z\right)}\Vert^2_2
\end{equation}
where $\beta$ is a weight coefficient, and $sg\left(\cdot\right)$ is the stop-gradient operator. The first term on the right-hand-side of Eq. \ref{eq:stage_1_loss} is the reconstruction loss, with $\Tilde{X}$ denoting the reconstructed samples, the second term is the codebook loss that updates the codebook by drawing the nearest code sample $Z$ closer to the embedding $Z_c$, and the third term is referred to as the 'commitment loss' \cite{van2017neural} that updates the encoder by moving the embedding towards the corresponding code.
The perceptual loss $\mathcal{L}_\text{percep} (X_,\Tilde{X})$ \cite{johnson2016perceptual} is computed using a pretrained VGG network \cite{simonyan2014very} to monitor the reconstruction error at multiple feature map levels. The GAN loss is defined as 

\begin{equation}\label{eq:gan_loss}
    \mathcal{L}_{GAN}=\log F_{\text{disc}}(X)+\log \left(1-F_{\text{disc}}(\Tilde{X})\right) 
\end{equation}
To implement $\mathcal{L}_{GAN} (X,\Tilde{X})$, a patched discriminator $F_{\text{disc}}$ is used to compute the KL-divergence between the ground truth $X$ and the reconstruction $\Tilde{X}$. With the loss functions $\mathcal{L}_{VQ}$, $\mathcal{L}_\text{percept}$, and $\mathcal{L}_{GAN}$, model training objective for Stage 1 can be formulated as solving the following optimization problem.
\begin{equation*}\label{eq:stage_1_goal}
    \operatorname*{min}_{\{E, \mathcal{C},D\}} \operatorname*{max}_{F_{\text{disc}}} \mathbb{E}_{X}\left[\mathcal{L}_{VQ}+\mathcal{L}_\text{percept}+\lambda\cdot\mathcal{L}_{GAN}\right]
\end{equation*}
where $\lambda$ is an adaptive weight \cite{esser2021taming} that balances the GAN loss and the other loss terms.

\subsection{Model Fine-Tuning for Data Completion (Stage 2)}
The goal of Stage-2 learning is to fine-tune the VQ-VAE model for data completion. During fine-tuning, the weights of decoder $D$ is kept constant while the codebook and the encoder weights are optimized using the same loss functions $\mathcal{L}_{VQ}$, $\mathcal{L}_\text{percept}$ and $\mathcal{L}_{GAN}$ from Equations \ref{eq:stage_1_loss} and \ref{eq:gan_loss}, respectively. 
An incomplete data sample $X_{\text{mask}}\in \mathcal{R}^{H\times W}$ is constructed by replacing the elements of $X$ in the masked out region $\mathcal{D}^{\text{mask}}$ with the value of zero. Numerically, such procedure can be implemented by computing $X_{\text{mask}}:=X\odot (\mathbf{1} - B)$, where $\odot$ denotes the element-wise multiplication of two matrices, and $B\in\{0,1\}^{H\times W}$ is a Boolean matrix defined as follows.
\begin{equation*}
    B(i, j) = 
  \begin{cases}
    0, &\text{if } X_{\text{mask}}(i,j) \text{ is a missing value,}\\
    1, &\text{if } X_{\text{mask}}(i,j) \text{ is a known value of 0,}
  \end{cases}
\end{equation*}
The model input for Stage-2 learning is obtained by stacking $X_{\text{mask}}\in \mathcal{R}^{H\times W}$, $B$ and $\mathbf{1}-B$ (where $\mathbf{1}$ denotes a $H\times W$ matrix of all $1$'s). The stacked input is introduced to distinguish between the case where an input collocation point has a value of zero because of zero reference value and the case where the zero value is a result of missing data. Concretely, the model fine-tuning objective for Stage 2 learning can be formulated as solving the following optimization problem.
\begin{equation}\label{eq:stage_2_goal}
    \operatorname*{min}_{\{E, \mathcal{C}\}} \operatorname*{max}_{F_{\text{disc}}} \mathbb{E}_{\{X,X_{\text{mask}}\}}\left[\mathcal{L}_{VQ}+\mathcal{L}_\text{percept}+\lambda\cdot\mathcal{L}_{GAN}\right]
\end{equation}
Since for data completion, only the masked region of the decoder output $\Tilde{X}$ is needed, the loss functions $\mathcal{L}_{VQ}$, $\mathcal{L}_\text{percept}$ and $\mathcal{L}_{GAN}$ in Formula \ref{eq:stage_2_goal} are evaluated only in the masked region $\mathcal{D^{\text{mask}}}$ by replacing $\Tilde{X}$ in Equations \ref{eq:stage_1_loss} and \ref{eq:gan_loss} with $(X_{\text{mask}}+\Tilde{X}\odot B)$.

\section{Experiments}
\subsection{Dataset}
The turbulent data for the training and test of the data completion network models is obtained from a numerical simulation of the 2-dimensional Kolmogorov flow \cite{chandler_2013_kolmogorov}, which solves the following vorticity equations with a forcing term, 
\begin{equation}
    \begin{aligned}
\frac{\omega (\mathbf{x},t)}{\partial t} + \left(\mathbf{u}(\mathbf{x},t) \cdot \nabla\right) \omega (\mathbf{x},t) &= \frac{1}{\textit{Re}} \nabla^2 \omega (\mathbf{x},t) + f(\mathbf{x}) , \quad \mathbf{x} \in (0,2\pi)^2,  t \in (0,T], \\
\nabla \cdot \mathbf{u}(\mathbf{x},t) &= 0, \quad  \mathbf{x} \in (0,2\pi)^2, t \in (0,T], \\
\omega (\mathbf{x},0) &= \omega_0(\mathbf{x}), \quad \mathbf{x} \in (0,2\pi)^2,
\end{aligned}
\label{kmflow-eq}
\end{equation}
where $\omega$ is the vorticity, $\mathbf{u}$ is the velocity vector, $\textit{Re}$ represents the Reynolds number set as $1000$ in the experiments, and $f(\mathbf{x})$ is the forcing term set to be $f(\mathbf{x}) = -4 \cos(4x_2) - 0.1 \omega(\mathbf{x}, t)$. $\omega, \mathbf{u}$ and $f$ are evaluated on the spatial coordinate $\mathbf{x}=[x_1, x_2]$ and time $t$. Equation \ref{kmflow-eq} is numerically solved under periodic boundary condition using a pseudo-spectral solver implemented in PyTorch \cite{pytorch} by Li \textit{et al.} \cite{PINO_li_2021}, with initial condition $\omega_0(\mathbf{x})$ generated from a Gaussian random field $\mathcal{N}(0, 7^{3/2}(-\Delta+49I)^{-5/2})$. A total of 40 simulation runs were collected from varying initial conditions on a $2048 \times 2048$ uniform grid, where 36 runs were used to obtain the training data and the remaining 4 runs were reserved for the test data. Each simulation run has a time duration of $10$ seconds. A spatial downsampling from $2048 \times 2048$ to $256 \times 256$ and a fixed time interval of $1/32$ second are applied to obtain the training and the test datasets. 

\begin{figure}[t]    
    \includegraphics[width=0.85\linewidth]{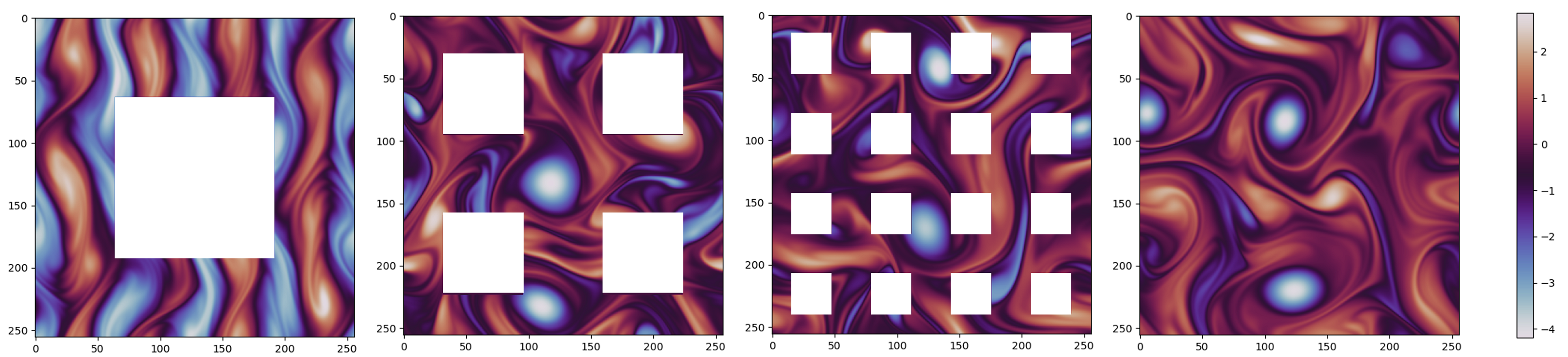}
    \captionof{figure}{Examples of 2D turbulent flow data considered in our data completion problem, where the first three examples from left to right are incomplete data samples with masked out regions used as model input and the right-most example is a complete data sample used as the ground truth reference. \label{fig:data-example}}
\end{figure}%
\subsection{Results}

We construct the incomplete data samples using one or multiple square masks to occlude vorticity values in the $256 \times 256$ 2D data domain. Three mask configurations were included in our data completion experiment. (1), A single mask of size $128 \times 128$ placed at the center of the data domain. (2), Four $64 \times 64$ masks spreading out evenly in the 2D space. (3), Sixteen $32 \times 32$ masks evenly distributed in the 2D space in a formation of four rows and four columns. A visualization of the three masks configurations are shown in Fig. \ref{fig:data-example}. The total area of masked out regions are the same among the three configurations. The main difference lies in the spatial span of continuous absence of data along the vertical or horizontal direction of the data domain. The motivation for choosing these configurations is to highlight how the scale of a continuous sub-domain of missing data - a key feature differentiating the data completion problem from the related data super-resolution problem - will affect the performance of the models. The location of the masks are fixed throughout model training and inference.

\begin{figure}[H]
    \includegraphics[width=0.9\linewidth]{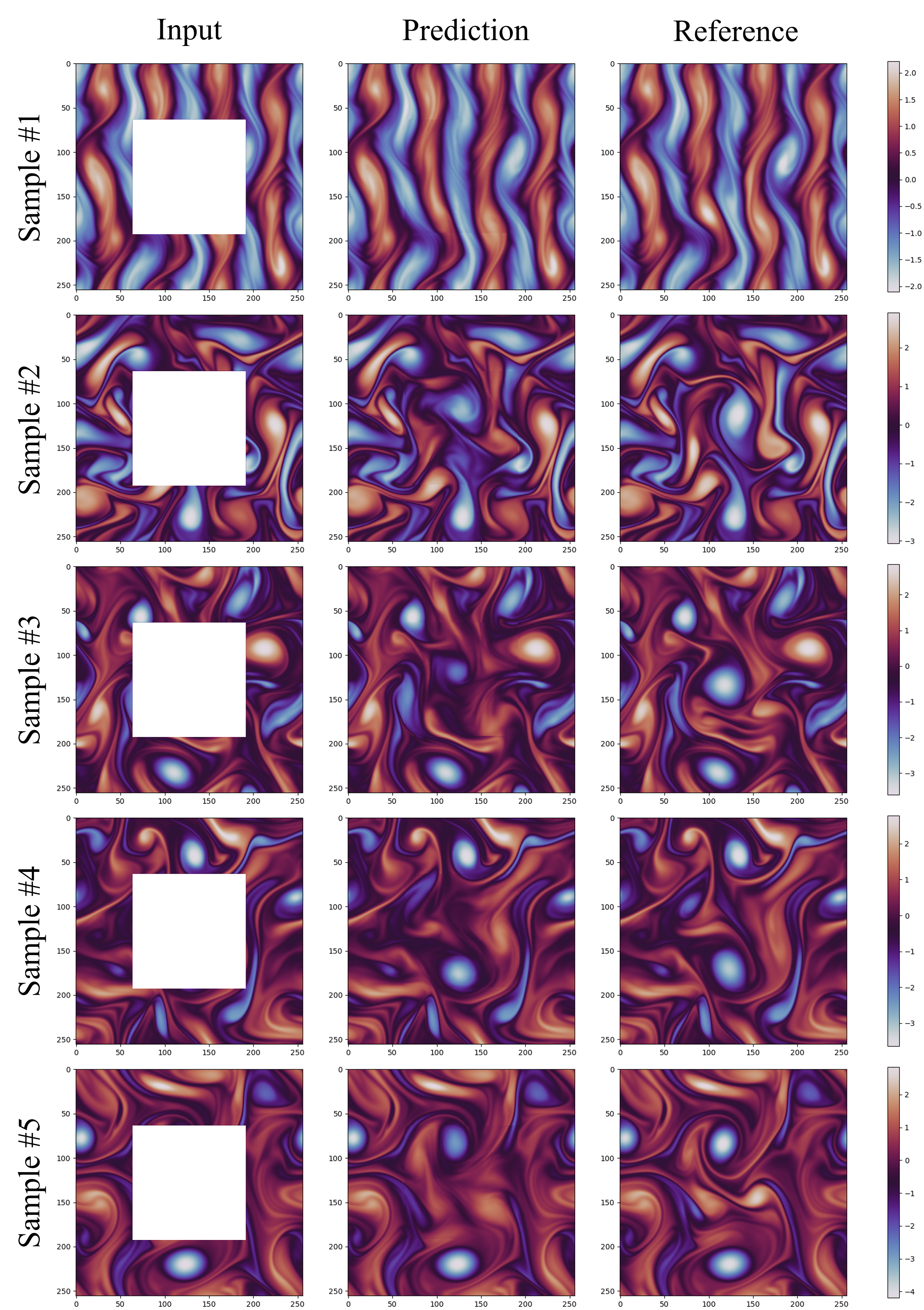}
    \captionof{figure}{Data completion samples from 1-mask experiment by proposed model. \label{fig:samples-mask1}}
    \end{figure}

\begin{figure}[H]
    \includegraphics[width=0.9\linewidth]{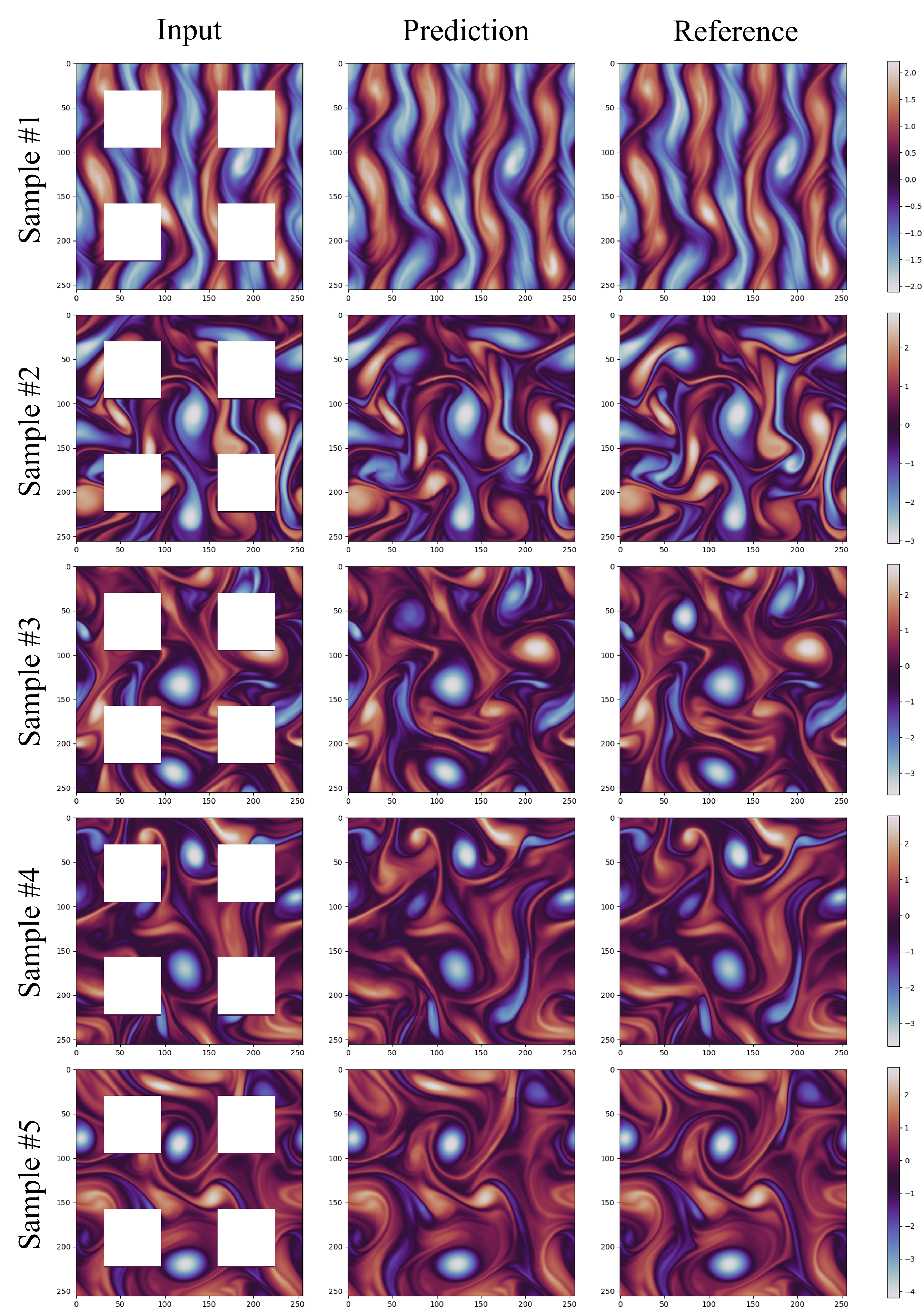}
    \captionof{figure}{Data completion samples from 4-mask experiment by proposed model. \label{fig:samples-mask4}}
    \end{figure}

\begin{figure}[H]
    \includegraphics[width=0.9\linewidth]{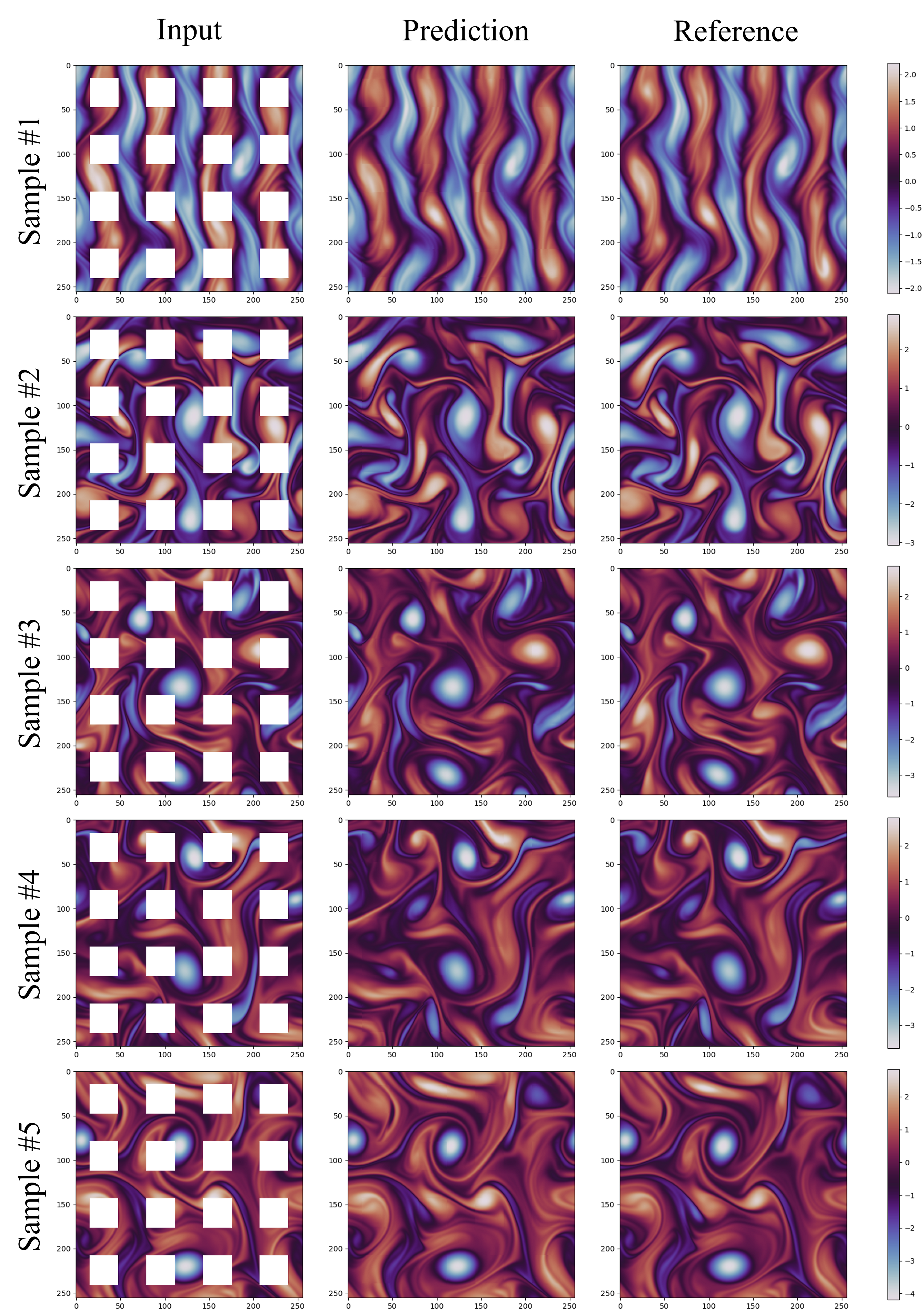}
    \captionof{figure}{Data completion samples from 16-mask experiment by proposed model. \label{fig:samples-mask16}}
\end{figure}

Figures \ref{fig:samples-mask1} , \ref{fig:samples-mask4} and \ref{fig:samples-mask16} provide visualization of five sampled results from data completion experiments. As can be observed, the 16-mask configuration produces the most accurate prediction of the missing data under all three mask configurations while the least accurate prediction comes from the 1-mask experiment. This comparison indicates that the extent of continuous data absence is a significant factor that affects the data completion quality of our model. A single mask of $256 \times 256$ poses a large challenge to the data completion task except for the first input sample, where a stripe-like pattern of laminar flows is dominant in the spatial distribution of vorticity values rather than the circular patterns of eddies commonly seen in the other four input samples. A possible reason for this observation is that the convolutional layers in the VQ-VAE allow the model to successfully recover the stripy patterns in the masked region using information from the unmasked region as a prior.

To quantify the performance of our model and the benchmark models, we adopted the relative $\mathcal{L}^{2}$ distance to calculate the loss on grid points from the masked region using the following equation:
\begin{equation}
    d_{\mathcal{L}^2}(\Tilde{X}, X) = \left[\frac{ \sum_{(i,j)\in \mathcal{D}^{\text{mask}}}\left(\Tilde{X}(i,j)-X(i,j)\right)^2 }{ \sum_{(i,j)\in \mathcal{D}^{\text{mask}}}\left(X(i,j)\right)^2 }\right]^{\frac{1}{2}}.
\label{eq:l2-error}
\end{equation}
The relative $\mathcal{L}^{2}$ distance in Eq. \ref{eq:l2-error} is introduced to evaluate the average point-wise prediction error of the data completion models. Under this metric, the completion accuracy of the three models from different mask configurations are summarized in Table \ref{tab:quant}. Compared with FNO and FactFormer, our model consistently yields a lower completion error. Among the three mask configurations, the 16-mask experiment produces the highest performance while the 1-mask experiment produces the lowest one. This is a result that aligns with the observations of the qualitative comparison shown by Figures \ref{fig:samples-mask1} , \ref{fig:samples-mask4} and \ref{fig:samples-mask16}. Similar trends can be observed from the two benchmark models, as their performances also reduce from 16-mask, 4-mask to 1-mask setting, with FactFormer generating a lower data completion error than FNO does. Results from Table \ref{tab:quant} and Figures \ref{fig:samples-mask1} , \ref{fig:samples-mask4} and \ref{fig:samples-mask16} suggest that, in data completion task, neural-network-based models are generally more vulnerable to extended continuous regions of missing data in turbulence flow, although such vulnerability can be mitigated when the fluid flow is less turbulent.

\begin{figure}[H]
    \includegraphics[width=\linewidth]{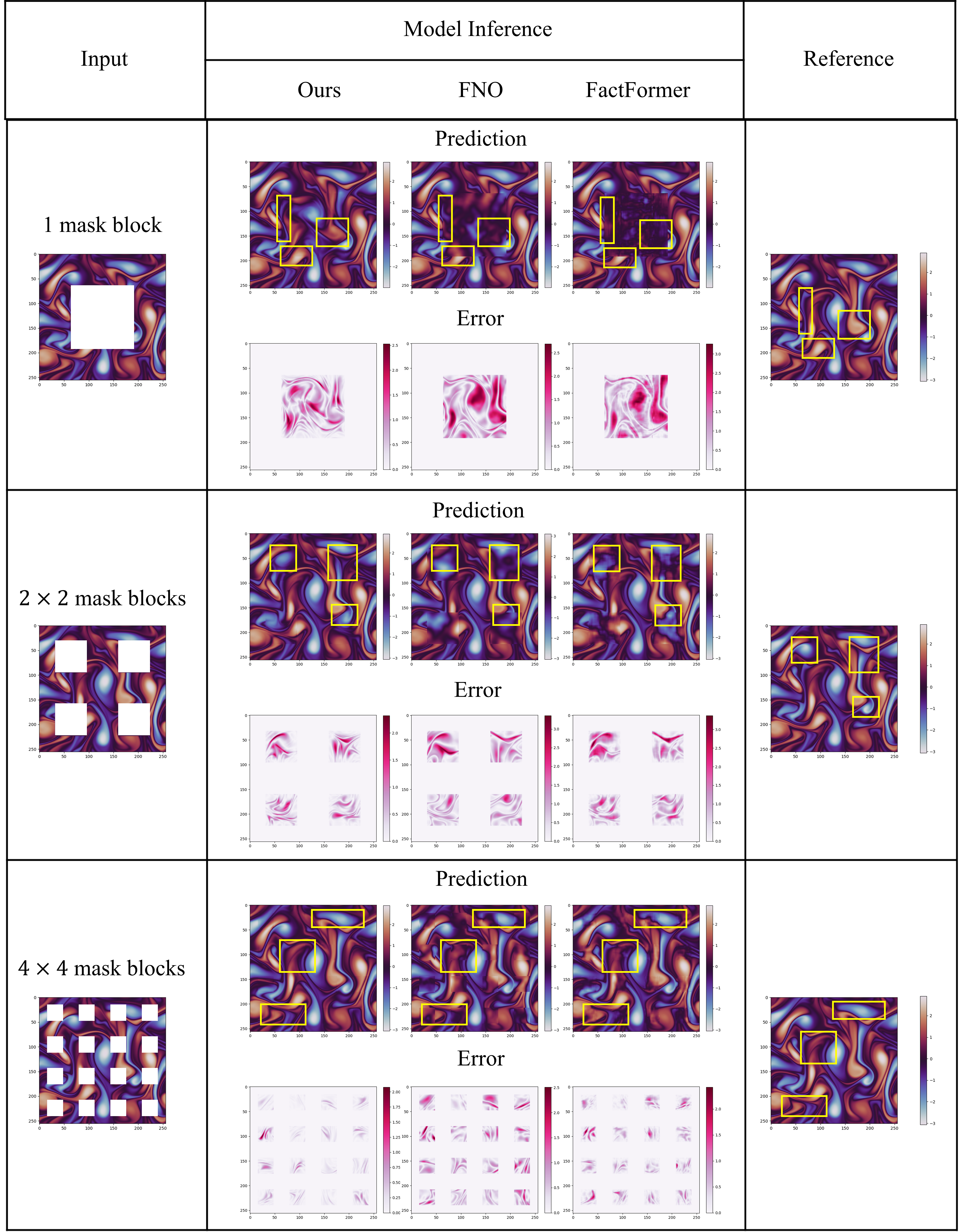}
    \captionof{figure}{Qualitative comparison of different data completion models on Sample \#2. \label{fig:model-comp}}
\end{figure}

\begin{figure}[H]
    \includegraphics[width=0.75\linewidth]{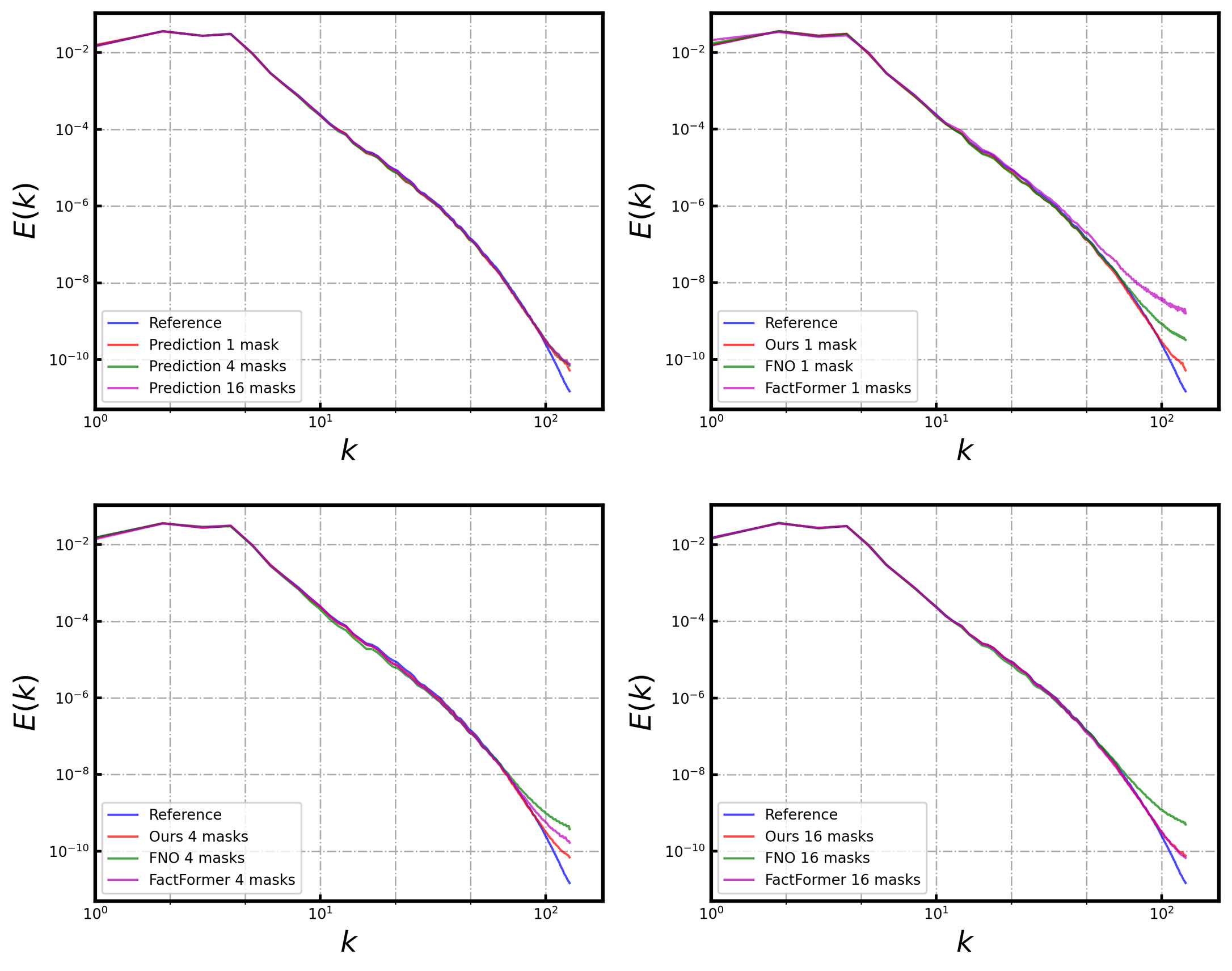}
    \captionof{figure}{Energy spectrum of data completion results from different models and mask configurations. \label{fig:energy-spectrum}}
\end{figure}

\begin{figure}[H]
    \includegraphics[width=0.65\linewidth]{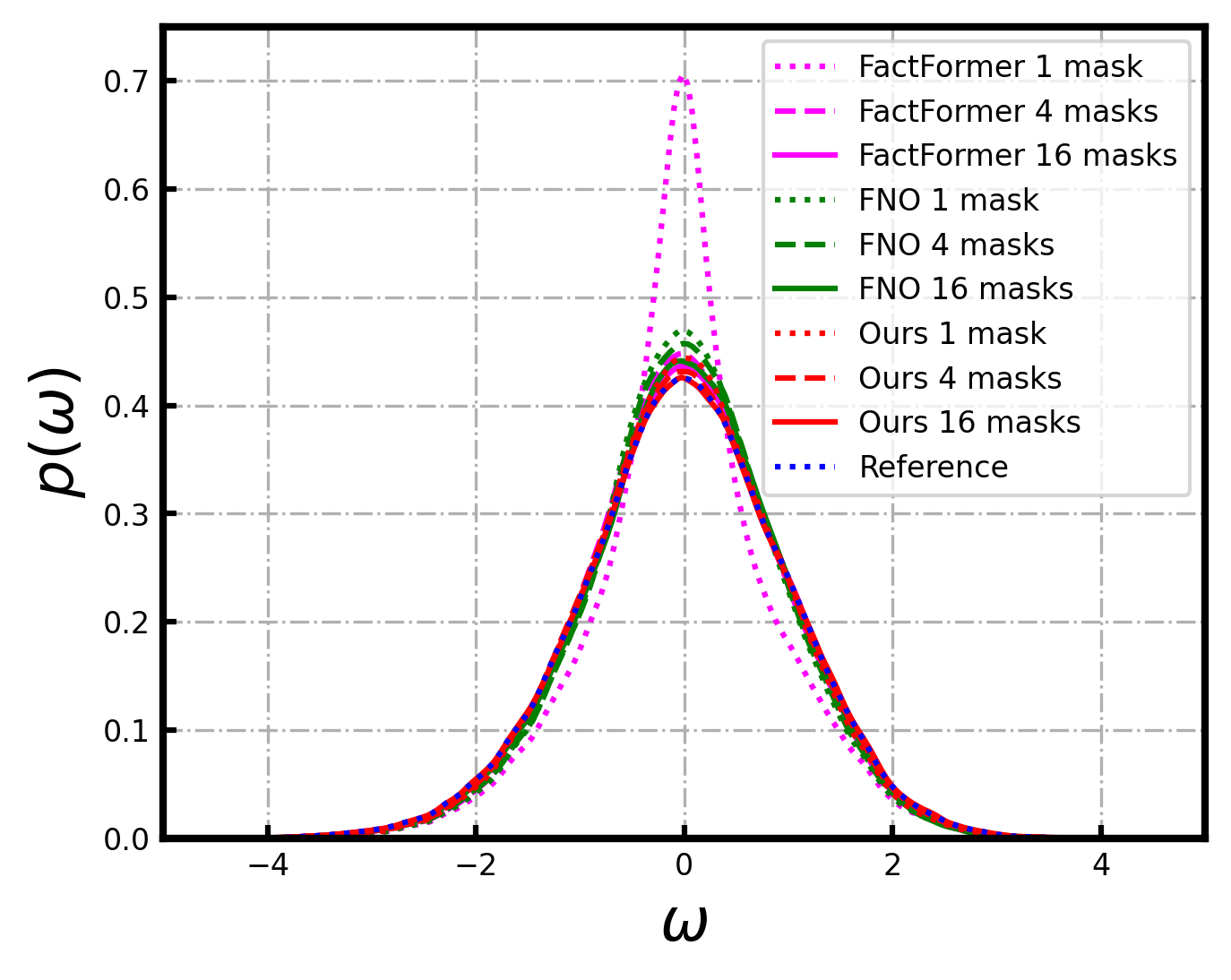}
    \captionof{figure}{One-dimensional vorticity distribution of data completion results from different models and mask configurations. \label{fig:vorticity-dist}}
\end{figure}

For a qualitative comparison between our model and the benchmark models, we present in Fig. \ref{fig:model-comp} a visualization of data completion results using Sample \#2 from Figures \ref{fig:samples-mask1} , \ref{fig:samples-mask4} and \ref{fig:samples-mask16}. 
Also included are visualization of the relative $\mathcal{L}^{2}$ error of data completion, and bounding boxes to highlight some details of data completion. Visualization from Fig. \ref{fig:model-comp} appears to be consistent with the quantitative evaluation shown in Table \ref{tab:quant}. 
\begin{table}
\centering
\begin{tabular}{|c|lll|}
\hline
\multicolumn{1}{|l|}{\multirow{2}{*}{\begin{tabular}[c]{@{}l@{}}Masks on\\ Input Data\end{tabular}}} & \multicolumn{3}{l|}{Completion Error in Relative $\mathcal{L}^2$ norm} \\ \cline{2-4} 
\multicolumn{1}{|l|}{}                                                                               & Ours       & FNO       & FactFormer      \\ \hline
16 masks                                                                                            & $\mathbf{0.1663}$          & $0.5175$          & $0.3374$               \\
4 masks                                                                                             & $\mathbf{0.3594}$          & $0.7321$          & $0.7044$               \\
1 mask                                                                                              & $\mathbf{0.6533}$          & $0.9278$          & $0.7134$               \\ \hline
\end{tabular}
\caption{Quantitative comparison of data completion results by different models. \label{tab:quant}
}
\end{table}
In an effort to offer additional perspectives for evaluating the authenticity of the data completion results, we plots the energy spectrum of the model predictions in Fig. \ref{fig:energy-spectrum}. 
As shown in the upper-left plot of Fig. \ref{fig:energy-spectrum}, data completion results from our model under all mask configurations are highly close to the ground truth reference, with the energy spectrum function value only starts to deviate from the reference after a wave number of $100$. Comparisons of our model and the benchmark models on energy spectrum function from 1-mask, 4-mask and 16-mask exmperiments are shown in the upper-right, lower-left and lower-right plots, respectively. In both the 1-mask and 4-mask configurations, the benchmark models deviate earlier from the reference as our model does, with FactFormer showing greater discrepancy than FNO in the 1-mask case and less discrepancy in the 4-mask case. For the 16-mask case, FactFormer yields a nearly identical energy spectrum as our model does. Another statistical property we use to compare the predictions of different models is the one-dimensional vorticity distribution shown in Fig. \ref{fig:vorticity-dist}, where the prediction of our model under the 16-mask configuration closely aligns with the reference, with an increased difference from the reference in the 4-mask configuration and a further-increased difference in the 1-mask configuration. Under each mask configuration, our model yields a closer vorticity distribution to the ground truth than the benchmark models do. The relative performance between FNO and FactFormer follows the same trend as shown by the energy spectrum plots. Figures \ref{fig:energy-spectrum} and \ref{fig:vorticity-dist} indicate that the data completion results by our model are closer to the reference ground truth in the statistical sense. 

\section{Conclusion}
This paper presents a deep learning method for data completion of 2D turbulent flow.  Our method employs a two-stage training procedure with a VQ-VAE model to predict vorticity values within a $256\times 256$ 2D space containing regions of missing data. Experimental findings demonstrate the superior performance of our model compared to two benchmark neural operator models across different mask configurations in terms of average point-wise prediction error and statistical properties such as the energy spectrum and vorticity distribution. Furthermore, by varying the mask configuration, we show that the deep neural network approaches studied in this work exhibit enhanced performance when faced with smaller areas of continuous data absence, yet struggle to maintain satisfying point-wise prediction accuracy in the presence of a single continuous region of the same total area. This observation implies the limitation of our proposed method in handling the ill-defined nature of turbulence data completion. Future directions to extent the current work includes the training strategy of progressive inpainting, optimization of the learned VQ latent space, investigating methods to generate preliminary predictions for masked regions, which could serve as valuable conditioning information for data completion tasks. One potential avenue involves leveraging a vanilla neural operator model to make predictions from initial conditions at a significantly lower resolution with a Gaussian smoothing step to regulate the complexity of turbulent data.

\section{Acknowledgement}
This work is supported by the Division of Chemical, Bioengineering, Environmental and Transport Systems at National Science Foundation and the start-up fund from the Department of Mechanical Engineering at Carnegie Mellon University, United States.

\newpage
\bibliography{ref}

\providecommand{\latin}[1]{#1}
\makeatletter
\providecommand{\doi}
  {\begingroup\let\do\@makeother\dospecials
  \catcode`\{=1 \catcode`\}=2 \doi@aux}
\providecommand{\doi@aux}[1]{\endgroup\texttt{#1}}
\makeatother
\providecommand*\mcitethebibliography{\thebibliography}
\csname @ifundefined\endcsname{endmcitethebibliography}  {\let\endmcitethebibliography\endthebibliography}{}
\begin{mcitethebibliography}{66}
\providecommand*\natexlab[1]{#1}
\providecommand*\mciteSetBstSublistMode[1]{}
\providecommand*\mciteSetBstMaxWidthForm[2]{}
\providecommand*\mciteBstWouldAddEndPuncttrue
  {\def\EndOfBibitem{\unskip.}}
\providecommand*\mciteBstWouldAddEndPunctfalse
  {\let\EndOfBibitem\relax}
\providecommand*\mciteSetBstMidEndSepPunct[3]{}
\providecommand*\mciteSetBstSublistLabelBeginEnd[3]{}
\providecommand*\EndOfBibitem{}
\mciteSetBstSublistMode{f}
\mciteSetBstMaxWidthForm{subitem}{(\alph{mcitesubitemcount})}
\mciteSetBstSublistLabelBeginEnd
  {\mcitemaxwidthsubitemform\space}
  {\relax}
  {\relax}

\bibitem[Habibi \latin{et~al.}(2021)Habibi, D'Souza, Dawson, and Arzani]{habibi2021integrating}
Habibi,~M.; D'Souza,~R.~M.; Dawson,~S.~T.; Arzani,~A. Integrating multi-fidelity blood flow data with reduced-order data assimilation. \emph{Computers in Biology and Medicine} \textbf{2021}, \emph{135}, 104566\relax
\mciteBstWouldAddEndPuncttrue
\mciteSetBstMidEndSepPunct{\mcitedefaultmidpunct}
{\mcitedefaultendpunct}{\mcitedefaultseppunct}\relax
\EndOfBibitem
\bibitem[Liu \latin{et~al.}(2022)Liu, Zhang, and Xia]{liu2022new}
Liu,~Y.; Zhang,~W.; Xia,~Z. A new data assimilation method of recovering turbulent mean flow field at high Reynolds numbers. \emph{Aerospace Science and Technology} \textbf{2022}, \emph{126}, 107328\relax
\mciteBstWouldAddEndPuncttrue
\mciteSetBstMidEndSepPunct{\mcitedefaultmidpunct}
{\mcitedefaultendpunct}{\mcitedefaultseppunct}\relax
\EndOfBibitem
\bibitem[Yousif \latin{et~al.}(2021)Yousif, Yu, and Lim]{yousif2021high}
Yousif,~M.~Z.; Yu,~L.; Lim,~H.-C. High-fidelity reconstruction of turbulent flow from spatially limited data using enhanced super-resolution generative adversarial network. \emph{Physics of Fluids} \textbf{2021}, \emph{33}\relax
\mciteBstWouldAddEndPuncttrue
\mciteSetBstMidEndSepPunct{\mcitedefaultmidpunct}
{\mcitedefaultendpunct}{\mcitedefaultseppunct}\relax
\EndOfBibitem
\bibitem[Fu \latin{et~al.}(2023)Fu, Helwig, and Ji]{fu2023semi}
Fu,~C.; Helwig,~J.; Ji,~S. Semi-Supervised Learning for High-Fidelity Fluid Flow Reconstruction. The Second Learning on Graphs Conference. 2023\relax
\mciteBstWouldAddEndPuncttrue
\mciteSetBstMidEndSepPunct{\mcitedefaultmidpunct}
{\mcitedefaultendpunct}{\mcitedefaultseppunct}\relax
\EndOfBibitem
\bibitem[Shu \latin{et~al.}(2023)Shu, Li, and Farimani]{shu2023physics}
Shu,~D.; Li,~Z.; Farimani,~A.~B. A physics-informed diffusion model for high-fidelity flow field reconstruction. \emph{Journal of Computational Physics} \textbf{2023}, \emph{478}, 111972\relax
\mciteBstWouldAddEndPuncttrue
\mciteSetBstMidEndSepPunct{\mcitedefaultmidpunct}
{\mcitedefaultendpunct}{\mcitedefaultseppunct}\relax
\EndOfBibitem
\bibitem[Zhu \latin{et~al.}(2020)Zhu, Park, Isola, and Efros]{zhu2020unpaired}
Zhu,~J.-Y.; Park,~T.; Isola,~P.; Efros,~A.~A. Unpaired Image-to-Image Translation using Cycle-Consistent Adversarial Networks. 2020\relax
\mciteBstWouldAddEndPuncttrue
\mciteSetBstMidEndSepPunct{\mcitedefaultmidpunct}
{\mcitedefaultendpunct}{\mcitedefaultseppunct}\relax
\EndOfBibitem
\bibitem[Goodfellow \latin{et~al.}(2014)Goodfellow, Pouget-Abadie, Mirza, Xu, Warde-Farley, Ozair, Courville, and Bengio]{goodfellow2014generative}
Goodfellow,~I.~J.; Pouget-Abadie,~J.; Mirza,~M.; Xu,~B.; Warde-Farley,~D.; Ozair,~S.; Courville,~A.; Bengio,~Y. Generative Adversarial Networks. 2014\relax
\mciteBstWouldAddEndPuncttrue
\mciteSetBstMidEndSepPunct{\mcitedefaultmidpunct}
{\mcitedefaultendpunct}{\mcitedefaultseppunct}\relax
\EndOfBibitem
\bibitem[Karras \latin{et~al.}(2019)Karras, Laine, and Aila]{karras2019stylebased}
Karras,~T.; Laine,~S.; Aila,~T. A Style-Based Generator Architecture for Generative Adversarial Networks. 2019\relax
\mciteBstWouldAddEndPuncttrue
\mciteSetBstMidEndSepPunct{\mcitedefaultmidpunct}
{\mcitedefaultendpunct}{\mcitedefaultseppunct}\relax
\EndOfBibitem
\bibitem[Ho \latin{et~al.}(2020)Ho, Jain, and Abbeel]{ho2020denoising}
Ho,~J.; Jain,~A.; Abbeel,~P. Denoising diffusion probabilistic models. \emph{Advances in Neural Information Processing Systems} \textbf{2020}, \emph{33}, 6840--6851\relax
\mciteBstWouldAddEndPuncttrue
\mciteSetBstMidEndSepPunct{\mcitedefaultmidpunct}
{\mcitedefaultendpunct}{\mcitedefaultseppunct}\relax
\EndOfBibitem
\bibitem[Li and Farimani(2020)Li, and Farimani]{li2020learning}
Li,~Z.; Farimani,~A.~B. Learning Lagrangian fluid dynamics with graph neural networks. \textbf{2020}, \relax
\mciteBstWouldAddEndPunctfalse
\mciteSetBstMidEndSepPunct{\mcitedefaultmidpunct}
{}{\mcitedefaultseppunct}\relax
\EndOfBibitem
\bibitem[Pant \latin{et~al.}(2021)Pant, Doshi, Bahl, and Barati~Farimani]{pant2021deep}
Pant,~P.; Doshi,~R.; Bahl,~P.; Barati~Farimani,~A. Deep learning for reduced order modelling and efficient temporal evolution of fluid simulations. \emph{Physics of Fluids} \textbf{2021}, \emph{33}, 107101\relax
\mciteBstWouldAddEndPuncttrue
\mciteSetBstMidEndSepPunct{\mcitedefaultmidpunct}
{\mcitedefaultendpunct}{\mcitedefaultseppunct}\relax
\EndOfBibitem
\bibitem[Howard \latin{et~al.}(2023)Howard, Perego, Karniadakis, and Stinis]{howard2023multifidelity}
Howard,~A.~A.; Perego,~M.; Karniadakis,~G.~E.; Stinis,~P. Multifidelity deep operator networks for data-driven and physics-informed problems. \emph{Journal of Computational Physics} \textbf{2023}, \emph{493}, 112462\relax
\mciteBstWouldAddEndPuncttrue
\mciteSetBstMidEndSepPunct{\mcitedefaultmidpunct}
{\mcitedefaultendpunct}{\mcitedefaultseppunct}\relax
\EndOfBibitem
\bibitem[Hemmasian and Farimani(2023)Hemmasian, and Farimani]{hemmasian2023multi}
Hemmasian,~A.; Farimani,~A.~B. Multi-scale Time-stepping of Partial Differential Equations with Transformers. \emph{arXiv preprint arXiv:2311.02225} \textbf{2023}, \relax
\mciteBstWouldAddEndPunctfalse
\mciteSetBstMidEndSepPunct{\mcitedefaultmidpunct}
{}{\mcitedefaultseppunct}\relax
\EndOfBibitem
\bibitem[Lorsung and Farimani(2024)Lorsung, and Farimani]{lorsung2024picl}
Lorsung,~C.; Farimani,~A.~B. PICL: Physics Informed Contrastive Learning for Partial Differential Equations. \emph{arXiv preprint arXiv:2401.16327} \textbf{2024}, \relax
\mciteBstWouldAddEndPunctfalse
\mciteSetBstMidEndSepPunct{\mcitedefaultmidpunct}
{}{\mcitedefaultseppunct}\relax
\EndOfBibitem
\bibitem[Lorsung \latin{et~al.}(2024)Lorsung, Li, and Barati~Farimani]{lorsung2024physics}
Lorsung,~C.; Li,~Z.; Barati~Farimani,~A. Physics Informed Token Transformer for Solving Partial Differential Equations. \emph{Machine Learning: Science and Technology} \textbf{2024}, \relax
\mciteBstWouldAddEndPunctfalse
\mciteSetBstMidEndSepPunct{\mcitedefaultmidpunct}
{}{\mcitedefaultseppunct}\relax
\EndOfBibitem
\bibitem[Fukami \latin{et~al.}(2019)Fukami, Fukagata, and Taira]{fukami2019super}
Fukami,~K.; Fukagata,~K.; Taira,~K. Super-resolution reconstruction of turbulent flows with machine learning. \emph{Journal of Fluid Mechanics} \textbf{2019}, \emph{870}, 106--120\relax
\mciteBstWouldAddEndPuncttrue
\mciteSetBstMidEndSepPunct{\mcitedefaultmidpunct}
{\mcitedefaultendpunct}{\mcitedefaultseppunct}\relax
\EndOfBibitem
\bibitem[Gao \latin{et~al.}(2021)Gao, Sun, and Wang]{gao2021}
Gao,~H.; Sun,~L.; Wang,~J.-X. Super-resolution and denoising of fluid flow using physics-informed convolutional neural networks without high-resolution labels. \emph{Physics of Fluids} \textbf{2021}, \emph{33}, 073603\relax
\mciteBstWouldAddEndPuncttrue
\mciteSetBstMidEndSepPunct{\mcitedefaultmidpunct}
{\mcitedefaultendpunct}{\mcitedefaultseppunct}\relax
\EndOfBibitem
\bibitem[Fukami \latin{et~al.}(2021)Fukami, Fukagata, and Taira]{fukami2021machine}
Fukami,~K.; Fukagata,~K.; Taira,~K. Machine-learning-based spatio-temporal super resolution reconstruction of turbulent flows. \emph{Journal of Fluid Mechanics} \textbf{2021}, \emph{909}\relax
\mciteBstWouldAddEndPuncttrue
\mciteSetBstMidEndSepPunct{\mcitedefaultmidpunct}
{\mcitedefaultendpunct}{\mcitedefaultseppunct}\relax
\EndOfBibitem
\bibitem[Kim \latin{et~al.}(2021)Kim, Kim, Won, and Lee]{kim2021unsupervised}
Kim,~H.; Kim,~J.; Won,~S.; Lee,~C. Unsupervised deep learning for super-resolution reconstruction of turbulence. \emph{Journal of Fluid Mechanics} \textbf{2021}, \emph{910}\relax
\mciteBstWouldAddEndPuncttrue
\mciteSetBstMidEndSepPunct{\mcitedefaultmidpunct}
{\mcitedefaultendpunct}{\mcitedefaultseppunct}\relax
\EndOfBibitem
\bibitem[Shu \latin{et~al.}(2023)Shu, Li, and Farimani]{Shu2023pi-diff}
Shu,~D.; Li,~Z.; Farimani,~A.~B. A physics-informed diffusion model for high-fidelity flow field reconstruction. \emph{Journal of Computational Physics} \textbf{2023}, \emph{478}, 111972\relax
\mciteBstWouldAddEndPuncttrue
\mciteSetBstMidEndSepPunct{\mcitedefaultmidpunct}
{\mcitedefaultendpunct}{\mcitedefaultseppunct}\relax
\EndOfBibitem
\bibitem[de~Avila Belbute-Peres \latin{et~al.}(2020)de~Avila Belbute-Peres, Economon, and Kolter]{belbuteperes2020combining}
de~Avila Belbute-Peres,~F.; Economon,~T.~D.; Kolter,~J.~Z. Combining Differentiable PDE Solvers and Graph Neural Networks for Fluid Flow Prediction. 2020\relax
\mciteBstWouldAddEndPuncttrue
\mciteSetBstMidEndSepPunct{\mcitedefaultmidpunct}
{\mcitedefaultendpunct}{\mcitedefaultseppunct}\relax
\EndOfBibitem
\bibitem[Li \latin{et~al.}(2021)Li, Li, and Farimani]{li2021tpu}
Li,~Z.; Li,~T.; Farimani,~A.~B. TPU-GAN: Learning temporal coherence from dynamic point cloud sequences. International Conference on Learning Representations. 2021\relax
\mciteBstWouldAddEndPuncttrue
\mciteSetBstMidEndSepPunct{\mcitedefaultmidpunct}
{\mcitedefaultendpunct}{\mcitedefaultseppunct}\relax
\EndOfBibitem
\bibitem[Xie \latin{et~al.}(2018)Xie, Franz, Chu, and Thuerey]{xie2018tempogan}
Xie,~Y.; Franz,~E.; Chu,~M.; Thuerey,~N. tempogan: A temporally coherent, volumetric gan for super-resolution fluid flow. \emph{ACM Transactions on Graphics (TOG)} \textbf{2018}, \emph{37}, 1--15\relax
\mciteBstWouldAddEndPuncttrue
\mciteSetBstMidEndSepPunct{\mcitedefaultmidpunct}
{\mcitedefaultendpunct}{\mcitedefaultseppunct}\relax
\EndOfBibitem
\bibitem[Esmaeilzadeh \latin{et~al.}(2020)Esmaeilzadeh, Azizzadenesheli, Kashinath, Mustafa, Tchelepi, Marcus, Prabhat, Anandkumar, \latin{et~al.} others]{esmaeilzadeh2020meshfreeflownet}
Esmaeilzadeh,~S.; Azizzadenesheli,~K.; Kashinath,~K.; Mustafa,~M.; Tchelepi,~H.~A.; Marcus,~P.; Prabhat,~M.; Anandkumar,~A.; others Meshfreeflownet: A physics-constrained deep continuous space-time super-resolution framework. SC20: International Conference for High Performance Computing, Networking, Storage and Analysis. 2020; pp 1--15\relax
\mciteBstWouldAddEndPuncttrue
\mciteSetBstMidEndSepPunct{\mcitedefaultmidpunct}
{\mcitedefaultendpunct}{\mcitedefaultseppunct}\relax
\EndOfBibitem
\bibitem[Ren \latin{et~al.}(2023)Ren, Rao, Liu, Ma, Wang, Wang, and Sun]{ren2023physr}
Ren,~P.; Rao,~C.; Liu,~Y.; Ma,~Z.; Wang,~Q.; Wang,~J.-X.; Sun,~H. PhySR: Physics-informed deep super-resolution for spatiotemporal data. \emph{Journal of Computational Physics} \textbf{2023}, \emph{492}, 112438\relax
\mciteBstWouldAddEndPuncttrue
\mciteSetBstMidEndSepPunct{\mcitedefaultmidpunct}
{\mcitedefaultendpunct}{\mcitedefaultseppunct}\relax
\EndOfBibitem
\bibitem[Um \latin{et~al.}(2020)Um, Brand, Fei, Holl, and Thuerey]{um2020solver}
Um,~K.; Brand,~R.; Fei,~Y.~R.; Holl,~P.; Thuerey,~N. Solver-in-the-loop: Learning from differentiable physics to interact with iterative pde-solvers. \emph{Advances in Neural Information Processing Systems} \textbf{2020}, \emph{33}, 6111--6122\relax
\mciteBstWouldAddEndPuncttrue
\mciteSetBstMidEndSepPunct{\mcitedefaultmidpunct}
{\mcitedefaultendpunct}{\mcitedefaultseppunct}\relax
\EndOfBibitem
\bibitem[Kochkov \latin{et~al.}(2021)Kochkov, Smith, Alieva, Wang, Brenner, and Hoyer]{kochkov2021machine}
Kochkov,~D.; Smith,~J.~A.; Alieva,~A.; Wang,~Q.; Brenner,~M.~P.; Hoyer,~S. Machine learning--accelerated computational fluid dynamics. \emph{Proceedings of the National Academy of Sciences} \textbf{2021}, \emph{118}, e2101784118\relax
\mciteBstWouldAddEndPuncttrue
\mciteSetBstMidEndSepPunct{\mcitedefaultmidpunct}
{\mcitedefaultendpunct}{\mcitedefaultseppunct}\relax
\EndOfBibitem
\bibitem[Sun \latin{et~al.}(2023)Sun, Yang, and Yoo]{sun2023neural}
Sun,~Z.; Yang,~Y.; Yoo,~S. A Neural PDE Solver with Temporal Stencil Modeling. 2023\relax
\mciteBstWouldAddEndPuncttrue
\mciteSetBstMidEndSepPunct{\mcitedefaultmidpunct}
{\mcitedefaultendpunct}{\mcitedefaultseppunct}\relax
\EndOfBibitem
\bibitem[Li \latin{et~al.}(2021)Li, Zheng, Kovachki, Jin, Chen, Liu, Azizzadenesheli, and Anandkumar]{PINO_li_2021}
Li,~Z.; Zheng,~H.; Kovachki,~N.; Jin,~D.; Chen,~H.; Liu,~B.; Azizzadenesheli,~K.; Anandkumar,~A. Physics-Informed Neural Operator for Learning Partial Differential Equations. 2021; \url{https://arxiv.org/abs/2111.03794}\relax
\mciteBstWouldAddEndPuncttrue
\mciteSetBstMidEndSepPunct{\mcitedefaultmidpunct}
{\mcitedefaultendpunct}{\mcitedefaultseppunct}\relax
\EndOfBibitem
\bibitem[Li \latin{et~al.}(2022)Li, Meidani, and Farimani]{li2022transformer}
Li,~Z.; Meidani,~K.; Farimani,~A.~B. Transformer for Partial Differential Equations' Operator Learning. \emph{arXiv preprint arXiv:2205.13671} \textbf{2022}, \relax
\mciteBstWouldAddEndPunctfalse
\mciteSetBstMidEndSepPunct{\mcitedefaultmidpunct}
{}{\mcitedefaultseppunct}\relax
\EndOfBibitem
\bibitem[Li and Farimani(2022)Li, and Farimani]{li2022graph}
Li,~Z.; Farimani,~A.~B. Graph neural network-accelerated Lagrangian fluid simulation. \emph{Computers \& Graphics} \textbf{2022}, \emph{103}, 201--211\relax
\mciteBstWouldAddEndPuncttrue
\mciteSetBstMidEndSepPunct{\mcitedefaultmidpunct}
{\mcitedefaultendpunct}{\mcitedefaultseppunct}\relax
\EndOfBibitem
\bibitem[Pant \latin{et~al.}(2021)Pant, Doshi, Bahl, and Barati~Farimani]{dl-rom2021}
Pant,~P.; Doshi,~R.; Bahl,~P.; Barati~Farimani,~A. {Deep learning for reduced order modelling and efficient temporal evolution of fluid simulations}. \emph{Physics of Fluids} \textbf{2021}, \emph{33}, 107101\relax
\mciteBstWouldAddEndPuncttrue
\mciteSetBstMidEndSepPunct{\mcitedefaultmidpunct}
{\mcitedefaultendpunct}{\mcitedefaultseppunct}\relax
\EndOfBibitem
\bibitem[Thuerey \latin{et~al.}(2020)Thuerey, Wei{\ss}enow, Prantl, and Hu]{Thuerey2020rans}
Thuerey,~N.; Wei{\ss}enow,~K.; Prantl,~L.; Hu,~X. Deep Learning Methods for Reynolds-Averaged Navier{\textendash}Stokes Simulations of Airfoil Flows. \emph{{AIAA} Journal} \textbf{2020}, \emph{58}, 25--36\relax
\mciteBstWouldAddEndPuncttrue
\mciteSetBstMidEndSepPunct{\mcitedefaultmidpunct}
{\mcitedefaultendpunct}{\mcitedefaultseppunct}\relax
\EndOfBibitem
\bibitem[Wiewel \latin{et~al.}(2019)Wiewel, Becher, and Thuerey]{latentphysics2021}
Wiewel,~S.; Becher,~M.; Thuerey,~N. Latent Space Physics: Towards Learning the Temporal Evolution of Fluid Flow. \emph{Computer Graphics Forum} \textbf{2019}, \emph{38}, 71--82\relax
\mciteBstWouldAddEndPuncttrue
\mciteSetBstMidEndSepPunct{\mcitedefaultmidpunct}
{\mcitedefaultendpunct}{\mcitedefaultseppunct}\relax
\EndOfBibitem
\bibitem[Sun \latin{et~al.}(2020)Sun, Gao, Pan, and Wang]{Sun2020surrogate}
Sun,~L.; Gao,~H.; Pan,~S.; Wang,~J.-X. Surrogate modeling for fluid flows based on physics-constrained deep learning without simulation data. \emph{Computer Methods in Applied Mechanics and Engineering} \textbf{2020}, \emph{361}, 112732\relax
\mciteBstWouldAddEndPuncttrue
\mciteSetBstMidEndSepPunct{\mcitedefaultmidpunct}
{\mcitedefaultendpunct}{\mcitedefaultseppunct}\relax
\EndOfBibitem
\bibitem[Brandstetter \latin{et~al.}(2022)Brandstetter, Worrall, and Welling]{mp-pde-solver}
Brandstetter,~J.; Worrall,~D.; Welling,~M. Message Passing Neural PDE Solvers. 2022; \url{https://arxiv.org/abs/2202.03376}\relax
\mciteBstWouldAddEndPuncttrue
\mciteSetBstMidEndSepPunct{\mcitedefaultmidpunct}
{\mcitedefaultendpunct}{\mcitedefaultseppunct}\relax
\EndOfBibitem
\bibitem[Pfaff \latin{et~al.}(2021)Pfaff, Fortunato, Sanchez-Gonzalez, and Battaglia]{pfaff2021learning}
Pfaff,~T.; Fortunato,~M.; Sanchez-Gonzalez,~A.; Battaglia,~P.~W. Learning Mesh-Based Simulation with Graph Networks. 2021\relax
\mciteBstWouldAddEndPuncttrue
\mciteSetBstMidEndSepPunct{\mcitedefaultmidpunct}
{\mcitedefaultendpunct}{\mcitedefaultseppunct}\relax
\EndOfBibitem
\bibitem[Jin \latin{et~al.}(2021)Jin, Cai, Li, and Karniadakis]{Jin2021nsfnet}
Jin,~X.; Cai,~S.; Li,~H.; Karniadakis,~G.~E. {NSFnets} (Navier-Stokes flow nets): Physics-informed neural networks for the incompressible Navier-Stokes equations. \emph{Journal of Computational Physics} \textbf{2021}, \emph{426}, 109951\relax
\mciteBstWouldAddEndPuncttrue
\mciteSetBstMidEndSepPunct{\mcitedefaultmidpunct}
{\mcitedefaultendpunct}{\mcitedefaultseppunct}\relax
\EndOfBibitem
\bibitem[Saini \latin{et~al.}(2016)Saini, Arndt, and Steinberg]{saini2016development}
Saini,~P.; Arndt,~C.~M.; Steinberg,~A.~M. Development and evaluation of gappy-POD as a data reconstruction technique for noisy PIV measurements in gas turbine combustors. \emph{Experiments in Fluids} \textbf{2016}, \emph{57}, 1--15\relax
\mciteBstWouldAddEndPuncttrue
\mciteSetBstMidEndSepPunct{\mcitedefaultmidpunct}
{\mcitedefaultendpunct}{\mcitedefaultseppunct}\relax
\EndOfBibitem
\bibitem[Venturi and Karniadakis(2004)Venturi, and Karniadakis]{venturi2004gappy}
Venturi,~D.; Karniadakis,~G.~E. Gappy data and reconstruction procedures for flow past a cylinder. \emph{Journal of Fluid Mechanics} \textbf{2004}, \emph{519}, 315--336\relax
\mciteBstWouldAddEndPuncttrue
\mciteSetBstMidEndSepPunct{\mcitedefaultmidpunct}
{\mcitedefaultendpunct}{\mcitedefaultseppunct}\relax
\EndOfBibitem
\bibitem[Pathak \latin{et~al.}(2016)Pathak, Krähenbühl, Donahue, Darrell, and Efros]{pathak2016inpaint}
Pathak,~D.; Krähenbühl,~P.; Donahue,~J.; Darrell,~T.; Efros,~A.~A. Context Encoders: Feature Learning by Inpainting. 2016 IEEE Conference on Computer Vision and Pattern Recognition (CVPR). 2016; pp 2536--2544\relax
\mciteBstWouldAddEndPuncttrue
\mciteSetBstMidEndSepPunct{\mcitedefaultmidpunct}
{\mcitedefaultendpunct}{\mcitedefaultseppunct}\relax
\EndOfBibitem
\bibitem[Song \latin{et~al.}(2016)Song, Yu, Zeng, Chang, Savva, and Funkhouser]{song2016semantic}
Song,~S.; Yu,~F.; Zeng,~A.; Chang,~A.~X.; Savva,~M.; Funkhouser,~T. Semantic Scene Completion from a Single Depth Image. 2016\relax
\mciteBstWouldAddEndPuncttrue
\mciteSetBstMidEndSepPunct{\mcitedefaultmidpunct}
{\mcitedefaultendpunct}{\mcitedefaultseppunct}\relax
\EndOfBibitem
\bibitem[Yu \latin{et~al.}(2018)Yu, Lin, Yang, Shen, Lu, and Huang]{Yu_2018_CVPR}
Yu,~J.; Lin,~Z.; Yang,~J.; Shen,~X.; Lu,~X.; Huang,~T.~S. Generative Image Inpainting With Contextual Attention. Proceedings of the IEEE Conference on Computer Vision and Pattern Recognition (CVPR). 2018\relax
\mciteBstWouldAddEndPuncttrue
\mciteSetBstMidEndSepPunct{\mcitedefaultmidpunct}
{\mcitedefaultendpunct}{\mcitedefaultseppunct}\relax
\EndOfBibitem
\bibitem[Buzzicotti \latin{et~al.}(2021)Buzzicotti, Bonaccorso, Leoni, and Biferale]{Buzzicotti2021inpaint}
Buzzicotti,~M.; Bonaccorso,~F.; Leoni,~P. C.~D.; Biferale,~L. Reconstruction of turbulent data with deep generative models for semantic inpainting from {TURB}-Rot database. \emph{Physical Review Fluids} \textbf{2021}, \emph{6}\relax
\mciteBstWouldAddEndPuncttrue
\mciteSetBstMidEndSepPunct{\mcitedefaultmidpunct}
{\mcitedefaultendpunct}{\mcitedefaultseppunct}\relax
\EndOfBibitem
\bibitem[Foucher \latin{et~al.}(2020)Foucher, Tang, da~Costa~de Azevedo, Kim, Gross, and Solenthaler]{foucher2020physicsaware}
Foucher,~S.; Tang,~J.; da~Costa~de Azevedo,~V.; Kim,~B.; Gross,~M.; Solenthaler,~B. Physics-Aware Flow Data Completion Using Neural Inpainting. 2020; \url{https://openreview.net/forum?id=BylldxBYwH}\relax
\mciteBstWouldAddEndPuncttrue
\mciteSetBstMidEndSepPunct{\mcitedefaultmidpunct}
{\mcitedefaultendpunct}{\mcitedefaultseppunct}\relax
\EndOfBibitem
\bibitem[Ronneberger \latin{et~al.}(2015)Ronneberger, Fischer, and Brox]{ronneberger2015unet}
Ronneberger,~O.; Fischer,~P.; Brox,~T. U-Net: Convolutional Networks for Biomedical Image Segmentation. 2015\relax
\mciteBstWouldAddEndPuncttrue
\mciteSetBstMidEndSepPunct{\mcitedefaultmidpunct}
{\mcitedefaultendpunct}{\mcitedefaultseppunct}\relax
\EndOfBibitem
\bibitem[Li \latin{et~al.}(2020)Li, Wang, Zhang, Du, and Tao]{li2020recurrent}
Li,~J.; Wang,~N.; Zhang,~L.; Du,~B.; Tao,~D. Recurrent feature reasoning for image inpainting. Proceedings of the IEEE/CVF conference on computer vision and pattern recognition. 2020; pp 7760--7768\relax
\mciteBstWouldAddEndPuncttrue
\mciteSetBstMidEndSepPunct{\mcitedefaultmidpunct}
{\mcitedefaultendpunct}{\mcitedefaultseppunct}\relax
\EndOfBibitem
\bibitem[Yang \latin{et~al.}(2020)Yang, Yang, Fu, Lu, and Guo]{yang2020learning}
Yang,~F.; Yang,~H.; Fu,~J.; Lu,~H.; Guo,~B. Learning texture transformer network for image super-resolution. Proceedings of the IEEE/CVF conference on computer vision and pattern recognition. 2020; pp 5791--5800\relax
\mciteBstWouldAddEndPuncttrue
\mciteSetBstMidEndSepPunct{\mcitedefaultmidpunct}
{\mcitedefaultendpunct}{\mcitedefaultseppunct}\relax
\EndOfBibitem
\bibitem[Johnson \latin{et~al.}(2016)Johnson, Alahi, and Fei-Fei]{johnson2016perceptual}
Johnson,~J.; Alahi,~A.; Fei-Fei,~L. Perceptual losses for real-time style transfer and super-resolution. Computer Vision--ECCV 2016: 14th European Conference, Amsterdam, The Netherlands, October 11-14, 2016, Proceedings, Part II 14. 2016; pp 694--711\relax
\mciteBstWouldAddEndPuncttrue
\mciteSetBstMidEndSepPunct{\mcitedefaultmidpunct}
{\mcitedefaultendpunct}{\mcitedefaultseppunct}\relax
\EndOfBibitem
\bibitem[Song \latin{et~al.}(2018)Song, Yang, Shen, Wang, Huang, and Kuo]{song2018spg}
Song,~Y.; Yang,~C.; Shen,~Y.; Wang,~P.; Huang,~Q.; Kuo,~C.-C.~J. Spg-net: Segmentation prediction and guidance network for image inpainting. \emph{arXiv preprint arXiv:1805.03356} \textbf{2018}, \relax
\mciteBstWouldAddEndPunctfalse
\mciteSetBstMidEndSepPunct{\mcitedefaultmidpunct}
{}{\mcitedefaultseppunct}\relax
\EndOfBibitem
\bibitem[Pan \latin{et~al.}(2021)Pan, Zhan, Dai, Lin, Loy, and Luo]{pan2021exploiting}
Pan,~X.; Zhan,~X.; Dai,~B.; Lin,~D.; Loy,~C.~C.; Luo,~P. Exploiting deep generative prior for versatile image restoration and manipulation. \emph{IEEE Transactions on Pattern Analysis and Machine Intelligence} \textbf{2021}, \emph{44}, 7474--7489\relax
\mciteBstWouldAddEndPuncttrue
\mciteSetBstMidEndSepPunct{\mcitedefaultmidpunct}
{\mcitedefaultendpunct}{\mcitedefaultseppunct}\relax
\EndOfBibitem
\bibitem[van~den Oord \latin{et~al.}(2018)van~den Oord, Vinyals, and Kavukcuoglu]{oord2018neural}
van~den Oord,~A.; Vinyals,~O.; Kavukcuoglu,~K. Neural Discrete Representation Learning. 2018\relax
\mciteBstWouldAddEndPuncttrue
\mciteSetBstMidEndSepPunct{\mcitedefaultmidpunct}
{\mcitedefaultendpunct}{\mcitedefaultseppunct}\relax
\EndOfBibitem
\bibitem[Kingma and Welling(2022)Kingma, and Welling]{kingma2022autoencoding}
Kingma,~D.~P.; Welling,~M. Auto-Encoding Variational Bayes. 2022\relax
\mciteBstWouldAddEndPuncttrue
\mciteSetBstMidEndSepPunct{\mcitedefaultmidpunct}
{\mcitedefaultendpunct}{\mcitedefaultseppunct}\relax
\EndOfBibitem
\bibitem[Razavi \latin{et~al.}(2019)Razavi, van~den Oord, and Vinyals]{razavi2019generating}
Razavi,~A.; van~den Oord,~A.; Vinyals,~O. Generating Diverse High-Fidelity Images with VQ-VAE-2. 2019\relax
\mciteBstWouldAddEndPuncttrue
\mciteSetBstMidEndSepPunct{\mcitedefaultmidpunct}
{\mcitedefaultendpunct}{\mcitedefaultseppunct}\relax
\EndOfBibitem
\bibitem[Li \latin{et~al.}(2020)Li, Kovachki, Azizzadenesheli, Bhattacharya, Stuart, Anandkumar, \latin{et~al.} others]{li2020fourier}
Li,~Z.; Kovachki,~N.~B.; Azizzadenesheli,~K.; Bhattacharya,~K.; Stuart,~A.; Anandkumar,~A.; others Fourier Neural Operator for Parametric Partial Differential Equations. International Conference on Learning Representations. 2020\relax
\mciteBstWouldAddEndPuncttrue
\mciteSetBstMidEndSepPunct{\mcitedefaultmidpunct}
{\mcitedefaultendpunct}{\mcitedefaultseppunct}\relax
\EndOfBibitem
\bibitem[Li \latin{et~al.}(2024)Li, Shu, and Barati~Farimani]{li2024scalable}
Li,~Z.; Shu,~D.; Barati~Farimani,~A. Scalable transformer for pde surrogate modeling. \emph{Advances in Neural Information Processing Systems} \textbf{2024}, \emph{36}\relax
\mciteBstWouldAddEndPuncttrue
\mciteSetBstMidEndSepPunct{\mcitedefaultmidpunct}
{\mcitedefaultendpunct}{\mcitedefaultseppunct}\relax
\EndOfBibitem
\bibitem[Cao(2021)]{cao2021choose}
Cao,~S. Choose a transformer: Fourier or galerkin. \emph{Advances in Neural Information Processing Systems} \textbf{2021}, \emph{34}, 24924--24940\relax
\mciteBstWouldAddEndPuncttrue
\mciteSetBstMidEndSepPunct{\mcitedefaultmidpunct}
{\mcitedefaultendpunct}{\mcitedefaultseppunct}\relax
\EndOfBibitem
\bibitem[Li \latin{et~al.}(2023)Li, Patil, Shu, and Farimani]{li2023latent}
Li,~Z.; Patil,~S.; Shu,~D.; Farimani,~A.~B. Latent Neural PDE Solver for Time-dependent Systems. NeurIPS 2023 AI for Science Workshop. 2023\relax
\mciteBstWouldAddEndPuncttrue
\mciteSetBstMidEndSepPunct{\mcitedefaultmidpunct}
{\mcitedefaultendpunct}{\mcitedefaultseppunct}\relax
\EndOfBibitem
\bibitem[Patil \latin{et~al.}(2023)Patil, Li, and Barati~Farimani]{patil2023hyena}
Patil,~S.; Li,~Z.; Barati~Farimani,~A. Hyena neural operator for partial differential equations. \emph{APL Machine Learning} \textbf{2023}, \emph{1}\relax
\mciteBstWouldAddEndPuncttrue
\mciteSetBstMidEndSepPunct{\mcitedefaultmidpunct}
{\mcitedefaultendpunct}{\mcitedefaultseppunct}\relax
\EndOfBibitem
\bibitem[Goswami \latin{et~al.}(2023)Goswami, Bora, Yu, and Karniadakis]{Goswami2023}
Goswami,~S.; Bora,~A.; Yu,~Y.; Karniadakis,~G.~E. In \emph{Machine Learning in Modeling and Simulation: Methods and Applications}; Rabczuk,~T., Bathe,~K.-J., Eds.; Springer International Publishing: Cham, 2023; pp 219--254\relax
\mciteBstWouldAddEndPuncttrue
\mciteSetBstMidEndSepPunct{\mcitedefaultmidpunct}
{\mcitedefaultendpunct}{\mcitedefaultseppunct}\relax
\EndOfBibitem
\bibitem[Esser \latin{et~al.}(2021)Esser, Rombach, and Ommer]{esser2021taming}
Esser,~P.; Rombach,~R.; Ommer,~B. Taming Transformers for High-Resolution Image Synthesis. 2021\relax
\mciteBstWouldAddEndPuncttrue
\mciteSetBstMidEndSepPunct{\mcitedefaultmidpunct}
{\mcitedefaultendpunct}{\mcitedefaultseppunct}\relax
\EndOfBibitem
\bibitem[Van Den~Oord \latin{et~al.}(2017)Van Den~Oord, Vinyals, \latin{et~al.} others]{van2017neural}
Van Den~Oord,~A.; Vinyals,~O.; others Neural discrete representation learning. \emph{Advances in neural information processing systems} \textbf{2017}, \emph{30}\relax
\mciteBstWouldAddEndPuncttrue
\mciteSetBstMidEndSepPunct{\mcitedefaultmidpunct}
{\mcitedefaultendpunct}{\mcitedefaultseppunct}\relax
\EndOfBibitem
\bibitem[Simonyan and Zisserman(2014)Simonyan, and Zisserman]{simonyan2014very}
Simonyan,~K.; Zisserman,~A. Very deep convolutional networks for large-scale image recognition. \emph{arXiv preprint arXiv:1409.1556} \textbf{2014}, \relax
\mciteBstWouldAddEndPunctfalse
\mciteSetBstMidEndSepPunct{\mcitedefaultmidpunct}
{}{\mcitedefaultseppunct}\relax
\EndOfBibitem
\bibitem[Chandler and Kerswell(2013)Chandler, and Kerswell]{chandler_2013_kolmogorov}
Chandler,~G.~J.; Kerswell,~R.~R. Invariant recurrent solutions embedded in a turbulent two-dimensional Kolmogorov flow. \emph{Journal of Fluid Mechanics} \textbf{2013}, \emph{722}, 554–595\relax
\mciteBstWouldAddEndPuncttrue
\mciteSetBstMidEndSepPunct{\mcitedefaultmidpunct}
{\mcitedefaultendpunct}{\mcitedefaultseppunct}\relax
\EndOfBibitem
\bibitem[Paszke \latin{et~al.}(2019)Paszke, Gross, Massa, Lerer, Bradbury, Chanan, Killeen, Lin, Gimelshein, Antiga, Desmaison, Köpf, Yang, DeVito, Raison, Tejani, Chilamkurthy, Steiner, Fang, Bai, and Chintala]{pytorch}
Paszke,~A. \latin{et~al.}  PyTorch: An Imperative Style, High-Performance Deep Learning Library. 2019; \url{https://arxiv.org/abs/1912.01703}\relax
\mciteBstWouldAddEndPuncttrue
\mciteSetBstMidEndSepPunct{\mcitedefaultmidpunct}
{\mcitedefaultendpunct}{\mcitedefaultseppunct}\relax
\EndOfBibitem
\end{mcitethebibliography}
\end{document}